\documentclass[letterpaper]{article} 

\newif\ifsubmission
\submissionfalse
\ifsubmission
  \usepackage[submission]{aaai2027} 
\else
  \usepackage[preprint]{aaai2027}   
\fi

\usepackage[hyphens]{url}  
\usepackage{graphicx} 
\usepackage{natbib}  
\usepackage{caption} 
\usepackage{amsmath}
\usepackage{amssymb}
\usepackage{amsfonts}
\usepackage{amsthm}

\usepackage{booktabs}
\usepackage{enumitem}
\usepackage{algorithm}
\usepackage{algpseudocode}
\usepackage{tabularx}
\usepackage{array}
\algrenewcommand\algorithmiccomment[1]{\hfill\texttt{\#}\,#1}
\DeclareMathOperator*{\argmax}{arg\,max}

\usepackage{newfloat}
\usepackage{listings}
\DeclareCaptionStyle{ruled}{labelfont=normalfont,labelsep=colon,strut=off} 
\floatstyle{ruled}
\newfloat{listing}{tb}{lst}{}
\floatname{listing}{Listing}

\newcommand{\where}{\textsc{where}}
\newcommand{\why}{\textsc{why}}

\title{HarnessBank: Semantic Gene-Bank Search with Gated Verification for Agent-Harness Self-Evolution}

\ifsubmission
  \author{Anonymous Submission}
  \affiliations{}
\else
  \author{
    Xiaotian Luo\textsuperscript{\rm 1,\rm 2},
    Dizhan Xue\textsuperscript{\rm 1,\rm 2},
    Fengxingyu Wang\textsuperscript{\rm 1,\rm 2},
    Chuanrui Hu\textsuperscript{\rm 1,\rm 2},
    Yafeng Deng\textsuperscript{\rm 1,\rm 2}\corresponding
  }
  \affiliations{
    \textsuperscript{\rm 1}EverMind AI \quad
    \textsuperscript{\rm 2}Shanda Group\\
    \{xiaotian.luo, dizhan.xue, ethan.wang\}@evermind.ai,
    \{chuanrui.hu, dengyafeng\}@shanda.com
  }
\fi

\begin{document}
\maketitle

\begin{abstract}
Large Language Models (LLMs) have enabled capable agents across diverse applications. Beyond the foundation model, the performance of an agent is governed by the surrounding \emph{agent harness}, including prompts, tools, control loops, etc. Automatically evolving this harness offers a promising pathway to agent improvement, yet existing approaches typically rely on greedy candidate selection and noisy self-generated feedback, rendering their gains susceptible to search collapse, task-specific overfitting, and poor verifiability.
To tackle these challenges, we introduce \textbf{HarnessBank}, a trustworthy agent-harness self-evolution framework that pairs a task agent with a separate evolver agent for iterative failure diagnosis, harness generation, and evolution verification. 
HarnessBank maintains a Harness Gene Bank composed of high-performing harnesses of different semantic coordinates. Those harnesses are reinvented, recombined, screened, and selected during the self-evolution procedure.
Moreover, we propose a Gated Harness Screening mechanism to efficiently filter high-quality harnesses and reduce the cost of evaluating numerous offspring harnesses.
Across seven agent benchmarks, HarnessBank produces consistent performance improvements from $5.1\%$ to $15.4\%$. Cross-model experiments further verify that the improvements come from the model-specific self-evolving process, instead of a universally optimal harness. 
Our code will be publicly available upon acceptance.
\end{abstract}

\section{Introduction}
\label{sec:intro}

Large Language Models (LLMs) have enabled increasingly capable autonomous agents that can reason, use tools, interact with environments, and complete complex tasks ranging from software engineering to web navigation~\citep{yao2023react,zhang2024codeagent,he2024webvoyager}. 
Beyond the foundation model, the empirical performance of an agent is fundamentally governed by the surrounding \emph{agent harness}~\citep{yang2024opro,khattab2024dspy,agrawal2026gepa}. 
The same frozen model can exhibit substantially different performance under different system prompts, injected knowledge, tool interfaces, control loops, recovery mechanisms, and runtime configurations. 
Unlike model weights, which may be closed, rented, or prohibitively expensive to update, the harness is often the only component that can be modified during deployment. 
Yet harnesses remain largely hand-engineered, making their development difficult to scale as models, tools, and application environments rapidly evolve. This motivates \emph{agent-harness self-evolution}~\citep{ren2026selfimprovementsmodernagenticsystems}, which aims to enable an agent to diagnose its own failures and iteratively improve its harness without updating the underlying model.

\begin{figure}[t]
  \centering
  \includegraphics[width=1\linewidth]{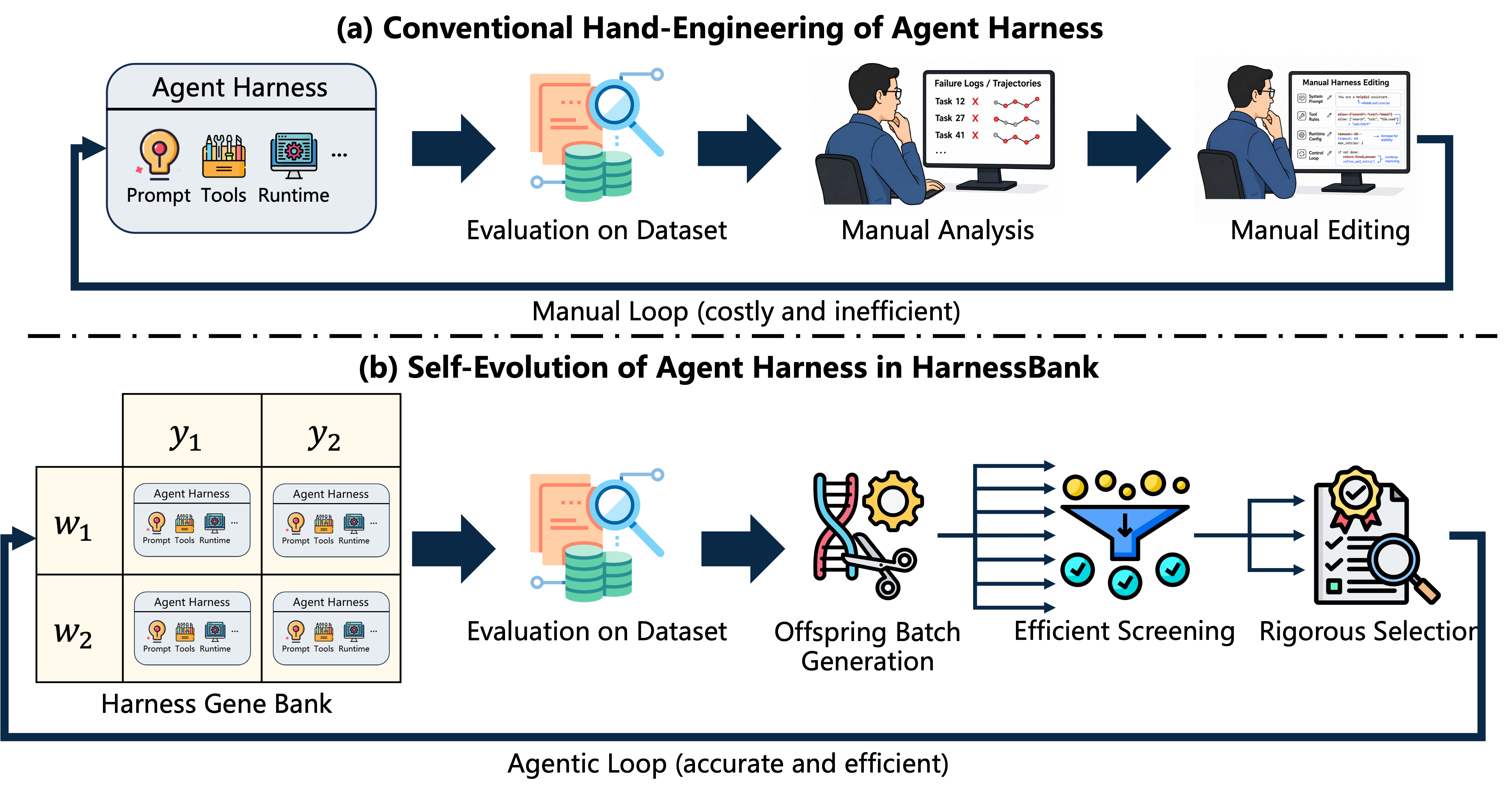}
  \caption{Conventional Hand-Engineering vs. Agent-Harness Self-Evolution: manual loop of agent harness development is costly and inefficient while self-evolution of agent harness in HarnessBank is accurate and efficient.}
  \label{fig:fig}
\end{figure}

Recent studies have begun to automate this process by allowing agents to inspect execution trajectories, propose harness modifications, and retain modifications that appear beneficial~\citep{zhang2026selfharness,chen2026harnessfix}. These studies demonstrate that harness-level adaptation can improve agent performance without model fine-tuning.
However, existing approaches commonly follow a relatively greedy evolution procedure in which a small number of locally promising edits are repeatedly promoted \citep{lin2026ahe,chen2026harnessx}.
Such procedures provide limited support for preserving structurally different solutions or combining complementary improvements, and may consequently collapse toward a narrow class of safe modifications, particularly prompt-level edits.
Moreover, for efficiency, candidates are often selected using single-run improvements, non-regression criteria, or model-predicted utility, without explicitly verifying whether the proposed mechanism is activated or whether the observed gain exceeds execution noise~\citep{ursekar2026vero,hambardzumyan2026aira2}.
These limitations make current self-evolution procedures susceptible to search collapse, task-specific overfitting, and unverifiable performance gains.

Based on the above observations, achieving trustworthy agent-harness self-evolution presents two fundamental challenges. 
\textbf{First, harness evolution requires structured exploration over a large and heterogeneous modification space.}
A harness may be improved through changes to prompts, injected knowledge, runtime control logic, recovery mechanisms, tool interfaces, or numerical configurations, yielding an effectively unbounded set of possible designs.
Greedy evolution tends to repeatedly exploit the current best candidate and gradually collapse toward conservative, incremental modifications after more ambitious offspring fail~\citep{ursekar2026vero}.
It may also retain patches that memorize particular training tasks rather than correct recurring failure mechanisms.
A successful self-evolution procedure must therefore preserve high-performing but semantically distinct harnesses, while enabling their useful mechanisms to be reinvented and recombined across generations.
\textbf{Second, identifying genuinely improved harnesses is both difficult and computationally expensive.}
Evaluating every offspring harness over the complete training set requires substantial agent rollouts, yet inexpensive evaluations on small task subsets can be unreliable.
Agent outcomes fluctuate across repeated executions, particularly on borderline tasks, while sandbox crashes, verifier timeouts, and other infrastructure artifacts may be incorrectly attributed to harness modifications.
Furthermore, an apparently successful patch may never activate during execution, or may overfit the tasks used to select it.
Trustworthy evolution therefore requires an efficient screening protocol that distinguishes functional and reproducible improvements from inactive mechanisms, stochastic variation, and evaluation artifacts.

To address these challenges, we introduce \textbf{HarnessBank}, a trustworthy agent-harness self-evolution framework that pairs a \emph{task agent} with a separate \emph{evolver agent}.
The task agent executes environment tasks under its current harness, whereas the evolver analyzes execution trajectories, diagnoses recurring failure mechanisms, and generates offspring harnesses without modifying the task model's parameters.
To prevent search collapse, HarnessBank introduces a \emph{Harness Gene Bank} that preserves high-performing harnesses across semantic coordinates.
Each bank cell is indexed by \emph{where} a harness modification acts and \emph{why} it is introduced, corresponding respectively to the modified harness component and the targeted failure pathology.
Harnesses targeting different failure mechanisms remain available for subsequent evolution.
The evolver can consequently produce offspring either by reinventing a mechanism from failure trajectories or by recombining compatible mechanisms preserved in different cells.
Quality-biased parent selection exploits the strongest discovered lineage, while the semantic bank retains complementary solutions that would otherwise be discarded by conventional greedy evolution.
To efficiently and reliably select offspring harnesses, HarnessBank further introduces \emph{Gated Harness Screening}.
Rather than immediately evaluating every offspring on the complete training set, HarnessBank first evaluates candidates on a sampled task subset and applies four sequential gates.
A validity gate excludes outcomes corrupted by infrastructure failures; an activation gate verifies that the proposed modification actually executes; a paired significance gate tests whether the observed gain is distinguishable from task-level variation; and a gain gate requires the candidate to outperform its parent.
Candidates that pass the screening procedure are subsequently evaluated on the complete training set before competing for admission to the Harness Gene Bank.
Across seven diverse agent benchmarks, HarnessBank consistently improves the performance of frozen task agents, with gains ranging from $5.1\%$ to $15.4\%$.

Our main contributions are summarized as follows:

\begin{itemize}[leftmargin=*]
\item We propose \textbf{HarnessBank}, an end-to-end framework for trustworthy agent-harness self-evolution. HarnessBank pairs a task agent with a separate evolver agent that iteratively diagnoses execution failures, generates candidate harnesses, and verifies their improvements without updating the underlying model parameters.

\item To mitigate search collapse and task-specific overfitting, we introduce a \textbf{Harness Gene Bank} that organizes high-performing harnesses using semantic coordinates. By preserving harnesses that address distinct failure pathologies, the gene bank supports both failure-guided reinvention and cross-cell recombination, enabling structured exploration beyond greedy optimization of a single harness lineage.

\item To efficiently and reliably identify improved harnesses, we propose \textbf{Gated Harness Screening}, which evaluates candidates on a sampled training subset through sequential checks of evaluation validity, mechanism activation, statistical significance, and performance gain. Only candidates that pass all gates undergo full training-set evaluation and compete for admission to the Harness Gene Bank.

\item  Experiments across seven diverse agent benchmarks demonstrate that HarnessBank consistently improves task agents, with performance gains ranging from $5.1\%$ to $15.4\%$. Cross-model experiments further show that the evolved harnesses are model-specific rather than universally optimal, indicating that the transferable capability lies in the failure-diagnosis, structured-search, and verification process.
\end{itemize}
\section{Related Work}

\subsection{LLM Agents and Agent Harnesses}

Large Language Models (LLMs) have progressed from passive text generators to the reasoning core of autonomous agents capable of solving long-horizon tasks~\citep{yao2023react,zhang2024codeagent,he2024webvoyager}. An \emph{LLM agent} places an LLM within an iterative perception--reasoning--action loop, allowing it to maintain task state, invoke external tools, observe environmental feedback, and revise its behavior over multiple steps. This system-level perspective has motivated extensive research on tool-augmented reasoning, memory-enhanced agents, workflow construction, and adaptive control~\citep{gao2026survey,xu2026lifeharness}.

\emph{Agent harness} denotes the executable scaffolding that surrounds a backbone LLM and converts it into a task-performing agent. It includes system and role prompts, injected knowledge and memory, tool interfaces, control-loop code, recovery and finalization policies, validators, and runtime configurations. Earlier work typically optimizes individual harness components in isolation, such as prompts, tools, memory, or workflows~\cite{hu-etal-2026-evermemos,su-etal-2026-u}. More recent studies have begun to treat the harness as a unified engineering surface whose heterogeneous components jointly determine agent behavior~\citep{gao2026survey,xu2026lifeharness}. This broader view is particularly important when model weights are frozen, closed, or costly to update.
However, manually developing a harness is often labor-intensive and error-prone. To address this, recent work has turned to automatic harness evolution, guided by the principle of improving agents with agents.

\subsection{Agent-Harness Self-Evolution}
A substantial body of agent-harness self-evolution can be understood as \emph{partial agent-harness self-evolution}, where execution feedback is used to improve one restricted component. Prompt-optimization methods automatically generate, refine, or select instructions~\citep{zhou2023ape,yang2024opro,khattab2024dspy,agrawal2026gepa}, while workflow-optimization approaches search over module compositions, operator orderings, and execution graphs~\citep{zhang2025aflow,lu2026empiricalmcts}. A parallel line updates contextual knowledge or memory from accumulated interactions~\citep{zhang2025ace,liu2026evolvemem,yu2026molem,zhang2026latentevolve}. 
These methods establish that individual harness components can be optimized from trajectories or task scores, but their search spaces are limited.

More recent work studies \emph{full agent-harness self-evolution}, which iteratively diagnoses execution trajectories, generates harness modifications, evaluates the resulting agent, and retains beneficial changes~\citep{lee2026metaharness,chen2026harnessx}. For example, Self-Harness clusters recurring failure patterns and proposes targeted harness edits under a non-regression rule~\citep{zhang2026selfharness}. HarnessFix maps trace-grounded failure diagnoses to scoped repair operators~\citep{chen2026harnessfix}. AHE exposes harness components as editable artifacts and validates modifications against self-declared behavioral predictions~\citep{lin2026ahe}. 

Despite this progress, existing methods often rely on greedy search and noisy selection criteria, making them vulnerable to search collapse, task-specific overfitting, and unreliable gains. HarnessBank addresses these issues with a \textbf{Harness Gene Bank} for preserving semantically diverse harnesses and \textbf{Gated Harness Screening} for efficiently verifying candidate improvements.
\begin{figure*}[t]
  \centering
  \includegraphics[width=0.9\textwidth]{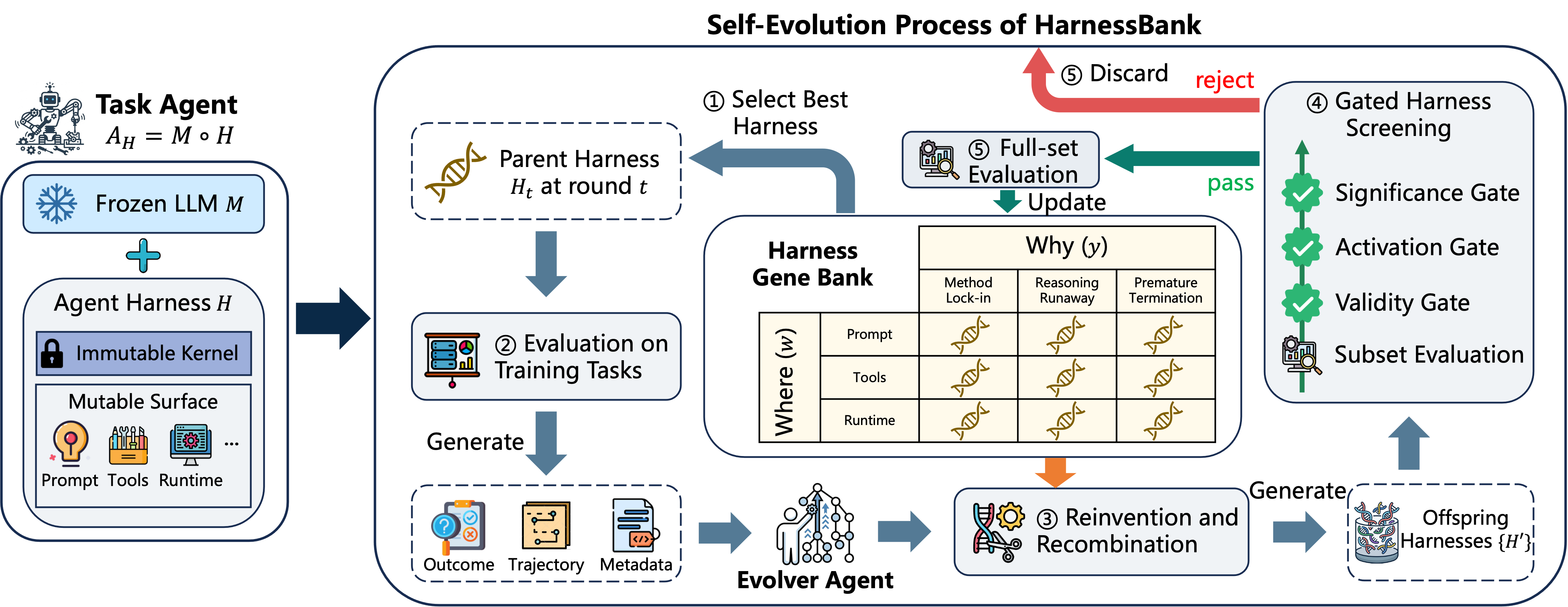}
  \caption{The framework of HarnessBank: 1) The best harness at round $t$ is selected; 2) The comprehensive diagnosis on training tasks is obtained; 3) The offspring harnesses are reinvented or recombined based on Harness Gene Bank; 4) High-quality harnesses are filtered by Gated Harness Screening; 5) Passed harnesses are evaluated and inserted into bank cells according to the principle of competitive selection.}
  \label{fig:loop}
\end{figure*}

\section{Method}
\label{sec:method}

As illustrated in Figure~\ref{fig:loop}, HarnessBank evolves the harness of a task agent through an iterative procedure.

\subsection{Problem Formulation}

\paragraph{Agent harness and mutable surface.}
Let $M$ denote a frozen LLM and $H$ its surrounding agent harness, including prompts, injected knowledge, tool specifications, runtime control logic, recovery mechanisms, and configurations.
The resulting task agent is:
\begin{equation}
A_H = M \circ H.
\end{equation}
Agent-harness self-evolution keeps the weights of $M$ fixed and modifies only $H$.
To preserve valid comparison, we partition the harness into an immutable kernel $\mathcal{K}$ and a mutable surface $\mathcal{X}$:
\begin{equation}
H=\mathcal{K}\cup\mathcal{X},
\quad
\mathcal{K}\cap\mathcal{X}=\varnothing.
\end{equation}
The kernel may contains evaluation, bookkeeping, self-evolution, and interface-critical code, whereas $\mathcal{X}$ contains the components that can be optimized.

\paragraph{Optimization objective.}
Let $\mathcal{D}_{\mathrm{tr}}$ denote the training tasks available during evolution and $\mathcal{D}_{\mathrm{te}}$ a test set withheld from all evolution decisions.
For task $i$, attempt $k$, and harness $H$, let $s_{i,k}(H)\in[0,1]$ be the task score.
We define the average utility score:
\begin{equation}
U(H;\mathcal{D})
=
\frac{1}{|\mathcal{D}|}
\sum_{i\in\mathcal{D}}
\frac{1}{K}
\sum_{k=1}^{K}s_{i,k}(H).
\end{equation}
Harness evolution searches for:
\begin{equation}
H^{*}
=
\argmax_{H\in\mathcal{H}(\mathcal{X})}
U(H;\mathcal{D}_{\mathrm{tr}}),
\end{equation}
where $\mathcal{H}(\mathcal{X})$ is the set of harnesses reachable through valid modifications to $\mathcal{X}$.
The test set is used only after evolution to assess the train-selected harness.

\subsection{Self-Evolution Based on Harness Gene Bank}
\label{sec:se}
HarnessBank separates four key components.
The \emph{task agent} executes tasks under the current harness. The \emph{evolver agent} diagnoses failures and proposes patches. The deterministic \emph{evaluator} controls sampling, scoring, activation logging, and statistical tests. The \emph{harness gene bank} preserves high-performing harnesses across semantic cells.

Conventional greedy evolution may collapse onto a narrow class of safe edits or over-specialize patches to individual training tasks.
To address this challenge, we introduce \emph{Harness Gene Bank} (HGB), which preserves high-performing candidate harnesses for iterative reinvention and recombination. HGB is inspired by the quality-diversity principle of MAP-Elites~\citep{mouret2015illuminating}.
Unlike conventional MAP-Elites, HGB uses semantic pathology descriptors, admits only verified candidates, and prioritizes the strongest lineage under a limited rollout budget.

\paragraph{Semantic cells.}
At round $t$, HGB maintains a categorical bank
\begin{equation}
\mathcal{A}_t:
\mathcal{W}\times\mathcal{Y}
\rightarrow
\mathcal{H}(\mathcal{X})\cup\{\varnothing\},
\end{equation}
where $\mathcal{H}(\mathcal{X})$ denotes the set of valid harnesses, and each cell
$c=(w,y)$ corresponds to a preserved harness.
Here, $w$ specifies \emph{where} it acts, and $y$ specifies \emph{why} it is introduced.
The \where{} coordinate belongs to:
\begin{equation}
\mathcal{W}
=
{
\texttt{prompt},
\texttt{knowledge},
\texttt{runtime},
\texttt{config}
},
\end{equation}
which is determined from the modified component.
The \why{} coordinate belongs to an evolving pathology set $\mathcal{Y}$ and is inferred from failure trajectories.
%
%
For example, a selective recovery mechanism for reasoning-budget exhaustion occupies
$(\texttt{runtime},\texttt{thinking-runaway})$.
%
%
At initialization, the gene bank is blank and $\mathcal{A}_0(c)=\varnothing$ for any $c$.

Harness indexes harnesses by targeted pathology rather than task-identity biases to search for reusable corrections.
Harnesses addressing the same hypothesized mechanism compete within one cell and only the best one is preserved, whereas harnesses targeting different pathologies remain available for recombination.
HGB preserves diverse, organized, and high-quality harnesses, which effectively expand the exploration space of offspring harness generation described in the following.

\paragraph{Heredity and variation.}
At round $t$, let $H_t$ denote the currently best harness sampled from the vanilla harness and the harness gene bank according to the quality-biased selection rule:
\begin{equation}
H_t
=
\argmax_{H \in
\left(\{H_0\}\cup\operatorname{Im}(\mathcal{A}_t)\right)
\setminus\{\varnothing\}}
U\!\left(H;\mathcal{D}_{\mathrm{tr}}\right).
\label{eq:bst}
\end{equation}
where $\operatorname{Im}(\mathcal{A}_t)$ is the image of $\mathcal{A}_t$, i.e., the set of all available harnesses in HGB.
$H_t$ serves as the parent harness in this round and has been evaluated on $\mathcal{D}_{\mathrm{tr}}$ while entering the bank.
For each training task, we have its outcome $s_{i,k}(H_t)$, trajectory $\tau_{i,k}(H_t)$, and evaluation metadata $e_{i,k}(H_t)$, forming a comprehensive diagnosis:
\begin{equation}
\mathcal{L}(H_t;\mathcal{D}_{\mathrm{tr}})
=
\left\{
\bigl(
s_{i,k}(H_t),
\tau_{i,k}(H_t),
e_{i,k}(H_t)
\bigr)
\right\}_{i\in\mathcal{D}_{\mathrm{tr}},\,k\in[K]}.
\label{eq:diag}
\end{equation}
Based on the diagnosis $\mathcal{L}(H_t;\mathcal{D}_{\mathrm{tr}})$ and the whole HGB, the evolver agent generates a batch of offspring harnesses $\{H'\}$ with a set $\{(w',y')\}$ indicating \emph{where} and \emph{why} these harnesses are evolved. Each new harness may be either newly invented or a combination of compatible harnesses from different bank cells.
For convenience, we denote the semantic descriptor $(w',y')$ of the harness $H'$ as $d(H')$.

\paragraph{Evolved harness selection.}
Given a batch of evolved harnesses, we need to find harnesses that actually improve upon the parent harness $H_t$.
Since inspecting and evaluating numerous harnesses on $\mathcal{D}_{\mathrm{tr}}$ is costly, we propose Gated Harness Screening to efficiently filter $N$ harnesses $B=\{H'_j\}_{j=1}^N$ likely to perform better than $H_t$, which will be introduced in detail in Section \ref{sec:vca}.
We evaluate every filtered harness $H'_j$ on the whole $\mathcal{D}_{\mathrm{tr}}$ and update its corresponding cells by the principle of competitive selection:
\begin{equation}
\mathcal{A}_{t+1}(d(H'_j))
=
\begin{cases}
H'_j,
&
\mathcal{A}_{t}(d(p)) = \varnothing,
\\[1.5mm]
H'_j,
&
U(H'_j;\mathcal{D}_{\mathrm{tr}})
>
U(\mathcal{A}_{t}(d(p));\mathcal{D}_{\mathrm{tr}}),
\\[1.5mm]
\mathcal{A}_{t}(d(p)),
&
\text{otherwise}.
\end{cases}
\label{eq:archive-update}
\end{equation}
Archive admission supports continued exploration but does not constitute final held-out credit.

The self-evolution of harnesses terminates after at most $R$ rounds or after $P$ consecutive rounds without a cell update.
Finally, the best harness is selected by Eq. \ref{eq:bst} and evaluated on the test tasks $\mathcal{D}_{\mathrm{te}}$.
We note that the test set is never used in the harness self-evolution.
%

\subsection{Gated Harness Screening}
\label{sec:vca}
The offspring harnesses $\{H'\}$ at round $t$ may contain various defects and may not improve upon the parent harness $H_t$.
However, evaluating all of $\{H'\}$ on the whole training set $\mathcal{D}_{\mathrm{tr}}$ is costly.
Therefore, we propose Gated Harness Screening, which efficiently evaluates $\{H'\}$ on a randomly sampled subset $\mathcal{D}_{\mathrm{sub}}$ of $\mathcal{D}_{\mathrm{tr}}$, and integrates three logic gates to mitigate the subset bias and filter out defective harnesses.
The computation of these gates is based on a comprehensive diagnosis on $\mathcal{D}_{\mathrm{sub}}$, defined in Eq. \ref{eq:diag}.

\paragraph{Validity and activation gates.}
The validity gate ensures that evaluation outcomes are obtained under a functioning evaluation environment.
Infrastructure failures, such as sandbox crashes or verifier timeouts, trigger a predefined repair-and-retry procedure rather than being immediately counted as agent failures.
A candidate $H'$ passes when its evaluation produces a complete protocol-valid ledger:
\begin{equation}
g_{\mathrm{valid}}(H';\mathcal{D}_{\mathrm{sub}})
=
\mathbb{I}
\left[
\mathcal{L}(H';\mathcal{D}_{\mathrm{sub}})
\text{ is protocol-valid}
\right].
\end{equation}

The activation gate verifies that the proposed mechanism actually executes.
The increment of each candidate harness (i.e., harness patches in $H'$ and not in $H_0$) declares an activation specification and emits a deterministic beacon when triggered.
Let $b_{i,k}(H')\in\{0,1\}$ indicate whether the increment of $H'$ activates on task $i$ and attempt $k$.
Then
\begin{equation}
g_{\mathrm{act}}(H';\mathcal{D}_{\mathrm{sub}})
=
\mathbb{I}
\left[
\sum_{i=1}^{|\mathcal{D}_{\mathrm{sub}}|}
\sum_{k=1}^{K}b_{i,k}(H')>0
\right].
\end{equation}
An evolved but never-triggered harness is treated as inert.
%

\paragraph{Significance gate.}
HarnessBank compares candidate and parent harnesses on identity tasks.
For the candidate $H'$, the task-level paired difference of the $i$th training task and total $K$ attempts is defined as:
\begin{equation}
\delta_i
=
\frac{1}{K}
\sum_{k=1}^{K}
\left[
s_{i,k}(H')-s_{i,k}(H_{\mathrm{t}})
\right].
\end{equation}
For training tasks, the estimated improvement and paired statistic are
\begin{equation}
\widehat{\Delta}
=
\frac{1}{|\mathcal{D}_{\mathrm{sub}}|}\sum_{i=1}^{|\mathcal{D}_{\mathrm{sub}}|}\delta_i,
\qquad
z
=
\frac{\widehat{\Delta}}
{\widehat{\sigma}_{\delta}/\sqrt{|\mathcal{D}_{\mathrm{sub}}|}},
\end{equation}
where $\widehat{\sigma}_{\delta}$ is the sample standard deviation of the paired differences.
The significance gate indicates whether the improvement is significant:
\begin{equation}
g_{\mathrm{sig}}(H';\mathcal{D}_{\mathrm{sub}})
=
\mathbb{I}
\left[\widehat{\Delta}>0
\right]
\times
\mathbb{I}
\left[z\geq1.96
\right]
.
\end{equation}
Here, $1.96$ is the $97.5$th percentile of the standard normal distribution, corresponding to the conventional two-sided $5\%$ significance level.
Therefore, the paired significance gate removes between-task difficulty variance and makes the decision independently auditable.
%

\paragraph{Harness screening.}
Finally, we compute the integrated screening gate $g_{\mathrm{scn}}$ as follows:
\begin{equation}
\begin{aligned}
&g_{\mathrm{scn}}(H';\mathcal{D}_{\mathrm{tr}})\\
=
&g_{\mathrm{valid}}
\bigl(H';\mathcal{D}_{\mathrm{tr}}\bigr)
\times
g_{\mathrm{act}}
\bigl(H';\mathcal{D}_{\mathrm{tr}}\bigr)
\times
g_{\mathrm{sig}}\bigl(H';\mathcal{D}_{\mathrm{tr}}\bigr)
\label{eq:navigation}
\end{aligned}
\end{equation}
To improve efficiency, we sequentially compute the value $g_{\mathrm{scn}}$ of each candidate $H'$ and select the first-$N$ passed harnesses.
These filtered harnesses further enter the evolved harness selection for a rigorous evaluation, as introduced in Section \ref{sec:se}.

\section{Experiments}
\label{sec:experiments}

\subsection{Experimental Setup}

We evaluate HarnessBank on seven domains with disjoint train/test splits: Terminal-Bench-2
(TB2)~\citep{merrill2026terminalbench}, five from EvoAgentBench (LiveCode, Omni-MATH,
BrowseComp+, GDPval, SWE-bench)~\citep{gao2026evoagentbench}, and
AppWorld~\citep{trivedi2024appworld}. All experiments evolve the harness only, with the
backbone frozen; by default \textbf{Qwen3.6-27B}. Claude Opus~4.8 serves as the evolver
agent. The primary metric is per-task success (Pass@1) averaged over $K{=}3$ attempts per
task. We compare against two state-of-the-art harness self-evolution methods,
GEPA~\citep{agrawal2026gepa} and DGM~\citep{zhang2026dgm}.
%

\begin{figure*}[t]
  \centering
  \IfFileExists{figures/fig-baselines.pdf}{\includegraphics[width=0.9\textwidth]{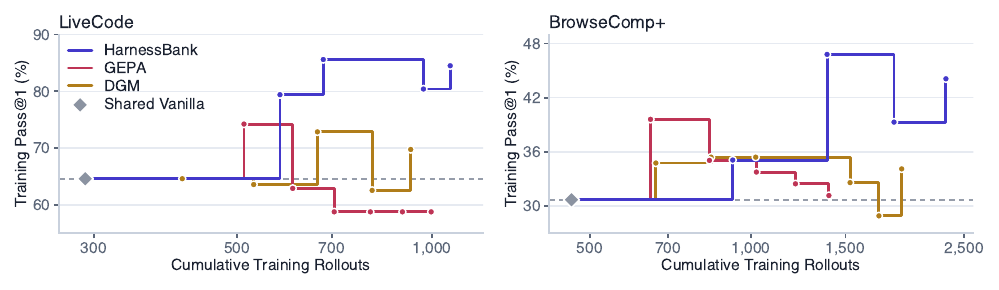}}{\figplaceholder{Fig.~\ref{fig:baselines}: export figures/fig-baselines.pdf}}
  \caption{Evolution trajectories (frozen Qwen3.6-27B): every harness produced, in order,
against cumulative rollouts (log scale), from a shared vanilla (diamond). No method climbs
monotonically --- GEPA's first candidate is its best on both domains. Training scores do
\emph{not} rank methods; ranking is decided once on the sealed test
(Figure~\ref{fig:testbars}).}
  \label{fig:baselines}
\end{figure*}

\subsection{Main results}

The train-selected harness improves over vanilla on \emph{every} test set, and every gain
clears the paired-$2\sigma$ bar (Table~\ref{tab:main}), from $+9$ to $+15.4$\%. The
recurring win is harness-level recovery of the frozen model's dominant failure mode (empty
``thinking-runaway'' turns), on several domains stacked with a verify-finalize self-check
(VF); our running example BrowseComp+ is one such recombination, credited at $+13.9$\%.

The seventh domain, SWE-bench, runs the same loop on a $101/26$ split: its harness is
strongly credited on training ($+13.9$\%, $n{=}101$) and lifts the test ($+5.1$\%), but
at $n{=}26$ even a real $+5$\% effect sits below the bar ($z{=}0.78$), so we report it as
\emph{preliminary} --- small-$n$ noise, not overfitting.

\begin{table*}[t]
\centering
\footnotesize
\setlength{\tabcolsep}{3pt}
\caption{Main results of HarnessBank on seven benchmarks. \emph{Vanilla} denotes the harness at initialization and \emph{Evolved} denotes the self-evolved harness by HarnessBank. Parentheses $=$ gain over vanilla (\%). Ret. $=$
test$\div$training gain. }
\label{tab:main}
\begin{tabular}{llccccccc}
\toprule
 & & \multicolumn{2}{c}{Training Pass@1} & \multicolumn{2}{c}{Test Pass@1} & & \multicolumn{2}{c}{Test Pass@3} \\
\cmidrule(lr){3-4}\cmidrule(lr){5-6}\cmidrule(lr){8-9}
Benchmark & Domain & Vanilla & Evolved & Vanilla & Evolved & Ret. & Vanilla & Evolved \\
\midrule
TB2               & Terminal Operation  & $37.7$ & $44.0$ ($+\phantom{0}6.3$) & $36.1$ & $\mathbf{45.4}$ ($+\phantom{0}9.3$) & $148\%$ & $52.8$ & $58.3$ ($+\phantom{0}5.5$) \\
LiveCode          & Code Generation     & $64.6$ & $85.6$ ($+21.0$) & $58.1$ & $\mathbf{71.8}$ ($+13.7$) & $65\%$ & $66.7$ & $79.5$ ($+12.8$) \\
Omni-MATH         & Math Reasoning      & $78.4$ & $91.1$ ($+12.7$) & $54.3$ & $\mathbf{66.0}$ ($+11.7$) & $92\%$ & $65.0$ & $75.0$ ($+10.0$) \\
BrowseComp+       & Web Research        & $30.7$ & $46.8$ ($+16.1$) & $16.9$ & $\mathbf{30.8}$ ($+13.9$) & $86\%$ & $33.8$ & $49.2$ ($+15.4$) \\
GDPval            & Knowledge Work      & $73.6$ & $82.0$ ($+\phantom{0}8.4$) & $43.7$ & $\mathbf{52.9}$ ($+\phantom{0}9.2$) & $110\%$ & $58.6$ & $67.2$ ($+\phantom{0}8.6$) \\
AppWorld          & App \& API Control  & $51.9$ & $69.7$ ($+17.8$) & $41.3$ & $\mathbf{56.7}$ ($+15.4$) & $86\%$ & $67.3$ & $75.6$ ($+\phantom{0}8.3$) \\
SWE-bench         & Repo-Level Bug Fix  & $55.4$ & $69.3$ ($+13.9$) & $47.4$ & $\mathbf{52.6}$ ($+\phantom{0}5.1$)          & $37\%$ & $61.5$ & $69.2$ ($+\phantom{0}7.7$) \\
\bottomrule
\end{tabular}
\end{table*}

\subsection{Comparison with Existing Methods}
\label{sec:baselines}

We compare HarnessBank against two state-of-the-art methods: GEPA~\citep{agrawal2026gepa}, which is
\emph{prompt-only}, and the Darwin G\"odel Machine (DGM)~\citep{zhang2026dgm}, which self-modifies
openly but \emph{ungated}. Both run under HarnessBank's protocol --- same frozen Qwen3.6-27B
as task agent \emph{and} proposer, same splits, same paired-$2\sigma$ ruler --- on comparable
budgets ($780$--$2{,}310$ rollouts each, within $2.1\times$ on any domain), with GEPA outspending
us on two domains and DGM on one (Figure~\ref{fig:baselines}).

\begin{figure}[t]
  \centering
  \IfFileExists{figures/fig-testbars.pdf}{\includegraphics[width=\columnwidth]{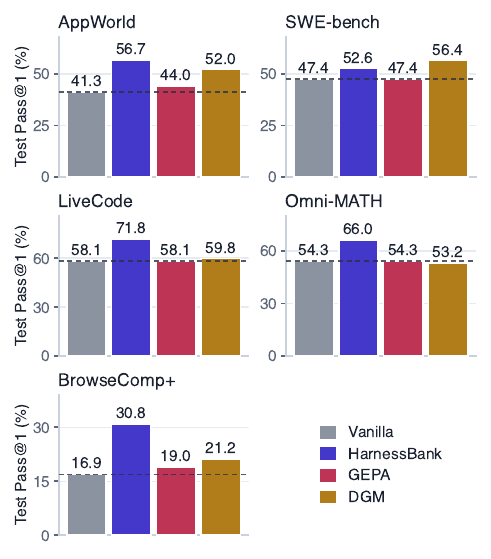}}{\figplaceholder{Fig.~\ref{fig:testbars}: export figures/fig-testbars.pdf}}
  \caption{Test Pass@1 ($K{=}3$) of different methods. Dashed lines denote the performance of the vanilla harness. }
  \label{fig:testbars}
\end{figure}

Across the five sealed tests of Figure~\ref{fig:testbars}, HarnessBank is credited on
\emph{four} (SWE-bench falls short only at $n{=}26$), DGM on \emph{one}, and GEPA on
\emph{none}.

GEPA finds \emph{no} variant beating its seed on LiveCode in $47$ iterations:
thinking-runaway is not prompt-addressable. Where its search returns no train-improving
candidate it ships vanilla, so three of its five cells sit at the baseline. On AppWorld,
whose win surface \emph{is} partly prompt-expressible, its training gain washes out on the
sealed test ($+2.8$, $z{=}0.97$).

DGM clears the bar once: AppWorld, $+10.7$\% ($z{=}3.00$). Its other cells show the cost
of selecting without a significance gate. On LiveCode it picks its best of $15$ generations
on a $15$-task $K{=}1$ spike ($0.733$, regressing to $0.533$ on re-evaluation) and lands
uncredited ($z{=}0.66$); on Omni-MATH the harness it ships is \emph{worse} than vanilla
($-1.1$\%) --- an ungated loop can deploy a regression.

The contrast is thus not search effort but what the edits can reach and what survives the
gates. On SWE-bench, the one domain where a baseline scores highest, neither is credited and
$n{=}26$ cannot separate them.

\subsection{Significance and Generalization}
\label{sec:sig}

All six test gains are \emph{credited}: each per-task paired difference clears
$z\!\ge\!1.96$ on tasks the evolving agent never saw ($p$ from $<10^{-4}$ to $0.033$;
AppWorld strongest at $z{=}6.44$, $n{=}168$). Each is a \emph{single} comparison (the
train-selected winner scored once on held-out tasks), so the many candidate comparisons
during evolution cannot inflate it. The gate is two-sided: on TB2 it culled a significantly
\textbf{negative} candidate ($-6.4$\%).

Test lift retains $65$--$148\%$ of train lift; none collapses. Two benchmarks exceed their
training gain and three retain $86$--$92\%$; LiveCode, whose training gain is the largest in
the suite, retains $65\%$. Retention is a ratio of two
noisy estimates that a small denominator inflates (GDPval's $110\%$ divides by $+8.4$\%,
LiveCode's $65\%$ by $+21.0$\%), so we do not treat it as a test: the load-bearing claim is
the credited held-out gain.

pass@3 also rises on \emph{every} credited domain ($+5.5$ to $+15.4$\%), so the harness
\emph{expands the set of solvable tasks}, not just the per-attempt hit rate; pass@1 remains
our primary metric.

\subsection{Diagnosis and Cost}

Two pathologies recur, each fixed by a distinct \textsc{runtime} mechanism:
\emph{thinking-runaway empty} turns, fixed by selective recovery, and
\emph{premature/unverified finalization}, fixed by a verify-finalize self-check. On five
domains a single pathology dominates $49$--$88\%$ of vanilla failures and the matched
mechanism is credited on the test, under both deterministic and judge-based verifiers ---
evidence of a \emph{true} failure mode of the frozen model rather than a scoring artifact.
The two domains with heterogeneous failure maps still yield credited and preliminary
patches.

\why{} is an LLM-assigned \emph{hypothesis}, not ground truth: on AppWorld the loop
misdiagnosed a capability limit as a knowledge gap, and the gate rejected the proposed patch
(its own target task $0/24 \to 0/24$, $p{=}1.0$). Because \why{} only steers which candidates
to try while credit comes solely from the gate, a wrong label costs at most a rejected
candidate, never a bad harness.

\subsection{Cross-Model Dissociation}

Thinking-runaway dominates four of six domains, but the loop is not re-fixing one
qwen3.6-27B quirk. The recurring win is a \emph{model} property (reasoning-budget overrun):
the credited mechanism disables thinking only \emph{after} a runaway, and on LiveCode neither
a $16\times$ token budget nor a blanket thinking-off toggle reproduces its gain. Cold-started
on other models the loop evolves \emph{their own} credited harnesses, each targeting that
model's distinct pathology (Table~\ref{tab:crossmodel}).

The patches follow a \emph{pathology$\to$patch matching law}: each model's dominant pathology
draws a patch that helps it and not models with a different one. On AppWorld the 27B fails by
empty ``engagement'' turns, fixed by verify-finalize, whereas the 397B and
\textbf{Gemini~3~Flash} make careless errors, fixed by a submit-verify checklist; Gemini
reproduces the careless$\to$checklist match ($+13.5$\%, $z{=}4.15$), extending the law
across \emph{families}. Off the diagonal the mismatched patch is near-zero on the Qwen pair
($+0.2$, $+1.2$); on Gemini verify-finalize still gains $+5.8$, a gradient rather than a
switch because it shares a verify step with the checklist.

Omni-MATH supplies the converse. The two Qwen generations \emph{share} the pathology
(thinking-runaway on long competition math) and the 27B-evolved stack transfers nearly
loss-free ($+11.7$ native, $+11.0$ transferred, both credited). Gemini reasons too
\emph{little} instead, so the transplanted recovery never fires ($-1.5$) while its matched
patch turns the \emph{same} lever the other way, to saturation ($+15.3$, credited). Turning
it the wrong way is not merely neutral but harmful: $-15.7$ when stacked on the evolved 397B
harness. The two families need one lever turned in opposite directions: a credited harness is
a correction fitted to the model, not a universally good setting.

\begin{table}[t]
\centering
\footnotesize
\setlength{\tabcolsep}{3.5pt}
\caption{Cross-model matching law: Test $\Delta$Pass@1 (\%). \textbf{Bold} $=$ matched
patch (paired-$2\sigma$). A/B $=$ VF\,/\,SV on AppWorld, 27B-stack\,/\,raise-reasoning on
Omni-MATH.}
\label{tab:crossmodel}
\begin{tabular}{lllcc}
\toprule
Domain & Model & Dominant Failure & A & B \\
\midrule
AppWorld  & 27B    & Empty Engagement  & $\mathbf{+15.4}$ & $+1.2$ \\
          & 397B   & Careless          & $+0.2$           & $\mathbf{+13.6}$ \\
          & Gemini & Careless          & $+5.8$           & $\mathbf{+13.5}$ \\
\midrule
Omni-MATH & 27B    & Thinks Too Much   & $\mathbf{+11.7}$ & $+1.7$ \\
          & 397B   & Runaway (Shared)  & $\mathbf{+11.0}$ & $+0.7$ \\
          & Gemini & Thinks Too Little & $-1.5$           & $\mathbf{+15.3}$ \\
\bottomrule
\end{tabular}

\end{table}

\subsection{Ablation Study}

Train selection is a \emph{lower bound} on what generalizes: on GDPval a variant ranked
below the winner on train scored \emph{highest} on test ($+11.5\%$ vs.\ $+9.2$\%) --- hence
crediting on held-out data.

Does the paired-$2\sigma$ gate matter, or would prior work's ``mean improves'' rules suffice?
Table~\ref{tab:gates} ablates it on TB2 along what gets deployed, what enters the archive,
and whether the loop can stop. Deployment is unchanged (train-argmax already picks VF), but
two noise mechanisms enter as elites --- one inert, its activation beacon never firing ---
and false elites then seed parents, spending future budget on noise. Termination is decisive:
on post-convergence rounds with neutral candidates, phantom progress appears in
$62$--$76\%$ of rounds under single-run or $K{=}3$-mean crediting, so the loop stops only at
the cap, whereas paired-$2\sigma$ stops at the $10$-round floor.

\begin{table}[t]
\centering
\footnotesize
\setlength{\tabcolsep}{3pt}
\caption{Ablation results on TB2. Rows below HarnessBank denote modified variants.
\emph{False elites} $=$ evolved harnesses whose gains in the test cannot be distinguished from
noise; $>20$ (cap) $=$ the stop rule is never met.}
\label{tab:gates}
\begin{tabular}{lccc}
\toprule
TB2 Configuration & Test Pass@1 & False Elites & Rounds \\
\midrule
\textbf{HarnessBank} ($K{=}3$, $2\sigma$) & $\mathbf{45.4}$ & $\mathbf{0}$ & $\mathbf{10.0}$ \\
w/o $2\sigma$                       & $\pm 0.0$ & $+2$ & $>20$ (cap) \\
w/o confirm $+$ $2\sigma$           & $-1.6$ & $+3$ & $>20$ (cap) \\
Vanilla                             & $-9.3$ & --- & --- \\
\bottomrule
\end{tabular}
\end{table}

Two signs the archive earns its keep. First, on most domains the credited harness
stacks mechanisms drawn from more than one cell, and it is the per-cell elite that
keeps the second mechanism alive long enough to be recombined at all. Second, accepted edits
span all four levers rather than prompts alone. The diversity axis is \emph{semantic} rather than
per-task by design: an archive keyed on tasks would preserve harnesses indexed by the very
tasks used to select them, which overfits by construction.

\section{Conclusion}

In this paper, we presented \textbf{HarnessBank}, a trustworthy framework for agent-harness self-evolution without updating the underlying model. HarnessBank combines a semantic Harness Gene Bank, which preserves and recombines high-performing harnesses across distinct failure pathologies, with Gated Harness Screening, which efficiently filters functional and statistically reliable improvements. Experiments across seven agent benchmarks demonstrate consistent gains of $5.1\%$ to $15.4\%$. Cross-model results further show that the evolved harnesses adapt to model-specific failure patterns, suggesting that robust agent improvement arises from the diagnosis--search--verification process rather than from a universally optimal harness.

\bibliography{refs}

\end{document}